\documentclass[conference]{IEEEtran}
\IEEEoverridecommandlockouts
\usepackage{cite}
\usepackage{amsmath,amssymb,amsfonts}
\usepackage{algorithmic}
\usepackage{graphicx}
\usepackage{physics}
\usepackage{upgreek}
\usepackage{textcomp}
\usepackage{bm}
\usepackage{xcolor}
\usepackage{amsmath}
\usepackage{tikz}
\DeclareMathOperator*{\argmin}{arg\,min}

\begin{document}

\title{\vspace{10mm}Upper Extremity Load Reduction for Lower Limb Exoskeleton Trajectory Generation Using Ankle Torque Minimization\\
\thanks{{Yik Ben Wong\textsuperscript{1}},{ Yawen Chen\textsuperscript{2}}, {Kam Fai Elvis Tsang\textsuperscript{3}}, and {Ling Shi\textsuperscript{5}} are with the Department of Electronic and Computer Engineering, Hong Kong University of Science and Technology, Hong Kong, China. Emails:\texttt{
\{ybwong, ychenga, kftsang, eesling\}@ust.hk}

 Winnie Suk Wai Leung\textsuperscript{4} is with the Division of Integrative Systems and Design, Hong Kong University of Science and Technology, Hong Kong, China. Emails:\texttt{
\{eewswleung\}@ust.hk}

This work is supported by a Hong Kong ITF Fund GHP/001/18SZ.}

}

\author{{Yik Ben Wong\textsuperscript{1}},{ Yawen Chen\textsuperscript{2}}, {Kam Fai Elvis Tsang\textsuperscript{3}},{ Winnie Suk Wai Leung\textsuperscript{4}}, and {Ling Shi\textsuperscript{5}}
}

\maketitle

\begin{abstract}
Recently, the lower limb exoskeletons  {which provide mobility for paraplegic patients to support their daily life have drawn much attention}. However, the pilots are required to apply excessive force through a pair of crutches to maintain balance during walking. This paper proposes a novel gait trajectory generation algorithm for exoskeleton locomotion on flat ground and stair which {aims} to minimize the force applied by the pilot without increasing the degree of freedom (DoF) of the system. First, the system is modelled as a {five-link} mechanism dynamically for torque computing. Then, an optimization approach is used to generate the trajectory minimizing the ankle torque which is correlated to the supporting force. Finally, experiment is conducted to compare the different gait generation {algorithms} through measurement of ground reaction force (GRF) applied on the crutches.
\end{abstract}


\section{Introduction}
The number of patients with locomotion disorder caused by stroke or spinal cord injury (SCI) is increasing over the years \cite{b1}. World Health Organization reported \cite{b12} that 250,000 - 500,000 patients are suffering from SCI around the globe every year. Even though wheelchair is an economical {alternative} solution for providing mobility, it exacerbates the medical consequences of paralysis such as the osteoporosis, muscle atrophy and pressure ulcers \cite{b2}. Studies show that exoskeletons are not only capable of improving both the muscle growth and bone marrow density, but also capable of facilitating social interaction at eye level \cite{b3}\cite{b4}. Young and Ferris \cite{b5} have reviewed the current state-of-the-art lower limb assistive exoskeleton system including HAL\cite{b7}, ReWalk\cite{b8}, Rex Bionics\cite{b9} and MINDWALKER\cite{b10} which apply force to the patients body with external actuators in order to provide mobility. 

Paraplegic patients encounter different environments such as slope, stair or obstacles in their daily {lives. Many} trajectory generation algorithms have been proposed to provide efficient and robust {implementation} to drive the exoskeleton. Some researchers have implemented the gait generation technology developed in bipedal humanoid on lower limb exoskeleton; for instance, linear inverted pendulum (LIP)\cite{b13} and zero moment point (ZMP)\cite{b14}. The {resulting} gait generated from the humanoid robot theory did not consider the human muscle dynamic and could be uncomfortable for the pilot.

Beside {the} gait generation method from bipedal robot theories, {some other} trajectory generation algorithms {use} healthy human gait as {a} reference to create human-like motion. Wu et al.\cite{b15} proposed an approach using autoencoder neural network to extract the gait spatial–temporal features of ground walking in different speed by multiple test objects. Gaussian process regression with automatic relevance determination is implemented to predict the feature by desired walking speed and user body parameters. Melo et al.\cite{b16} proposed a method using Fourier {decomposition to} extract the frequency components from the gait. Principal component analysis is then used to extract the feature of the frequency component and the trajectory is generated by reprojection based on ground-truth gait and body parameters. Other than level-ground walking, Xu et al \cite{b17} proposed an adaptive trajectory planning algorithm for stair-ascending and stair-descending using stair size and body parameters as inputs.  Infrared sensor and ZMP is installed to detect the size of the stair and to ensure safety during translation of standing leg and swing leg, respectively. The {aforementioned method} can generate a human-like gait, but the balance during the swing phase is not {considered}.  Chen et al.\cite{b18} proposed an online gait modification method for pre-designed gait to control the centre of pressure (CoP) of the system in order to increase the stability. The modification would trigger when the CoP {exceeds} the tolerable range to achieve active balancing, and the result is evaluated by measuring GRF {through} the crutches to verify whether the modification method is effective.     

Most of the exoskeleton systems require the pilot to balance through crutches, and if the pilots have to handle the high load caused by balancing frequently, the pilot would likely suffer serious shoulder pain\cite{b6}. The GRF of the crutches is a key index to assess the effectiveness of an exoskeleton. In other words, crutches GRF minimization would lead to better user experience of the exoskeleton and improved stability\cite{b6}{,} \cite{b18}.

In this paper, a 6 DoFs exoskeleton system is considered with 4 active DoFs on knee and hip joints and 2 semi-passive DoFs on ankle joints. A novel trajectory generation method is proposed to minimize the crutches GRF using an optimization approach, which {utilizes} the ankle torque as {an} approximation of the GRF from crutches and minimizes it dynamically in the swing phase for trajectory planning (shown in Fig. \ref{explain}), and the measurement of the GRF is used to evaluate the exoskeleton gait performance.

The motivation of this work is stated as follow.
\begin{enumerate}
  \item A generic exoskeleton gait generation algorithm would allow the pilot to {tackle} different environments. 

  \item Limited research have focused on model of crutches GRF.
  
  \item Limited research have considered reducing the GRF through gait trajectory generation.  
\end{enumerate}

The main contributions of this paper are summarized as follow.

\begin{enumerate}

  \item {We proposed a \textbf{novel} trajectory generation method which allows {a pilot} to walk through flat ground and stairs.}
  
  \item {We \textbf{demonstrated} the relationship between GRF of crutches and ankle torque of the exoskeleton.}
  
  \item {We proposed an optimization approach on the torque of the ankle joint in the swing phase {which} \textbf{significantly {reduces} the GRF on the crutches}.}

\end{enumerate}

The structure of this paper is organized as follow. {The test platform} is described in {Section} II, then the kinematic and dynamic model of the system is presented in  {Section}  III. In  {Section}  IV, the {details} of the optimization problem and the cost function are stated. The crutches GRF result {compared with} other trajectory generation methods is shown to the effectiveness of the proposed method in  {Section}  V. Finally, some of the limitation and future improvement are discussed in {Section} VI.

\section{Lower limb exoskeleton test platform}
The test platform is a wearable lower limb exoskeleton prototype developed by Xeno Dynamics Co., Ltd which aims to provide mobility to paraplegic patients. It is mainly constructed with aluminium and weighs 17 kg excluding the crutches and battery. It has in total 6 DoFs, among which 4 active DoFs are driven by brushless motors with harmonic gear reduction in hip flexion/extension and knee flexion/extension, and 2 semi-passive DoFs are provided by compression spring in ankle dorsiflexion/plantarflexion. In order to ensure the safety of the pilot, mechanical limit is designed to guarantee that the joint angles are in human joint range (more details are presented in Table \ref{tab1}). High resolution encoder and force-sensing resistor (FSR) are implemented in the sensing system for control and evaluation purpose. Encoders are installed at hip, knee and ankle joints to acquire angular information and FSR is placed under the crutches and feet of exoskeleton to obtain force {data}.   

\begin{table}[htbp]
\caption{Joint limit}
\begin{center}
\begin{tabular}{lclcl}
\hline
                      & Range (degree) &  \\ \hline
Hip Flexion/Extension & 100/80&  \\
Knee Flexion/Extension           &  100/0            &  \\
Ankle dorsiflexion/plantarflexion     & 20/0              &  \\\hline
\end{tabular}
\label{tab1}
\end{center}

\end{table}
\vspace{-4mm}
\begin{figure}[!htb]
\centerline{\includegraphics[scale = 0.26]{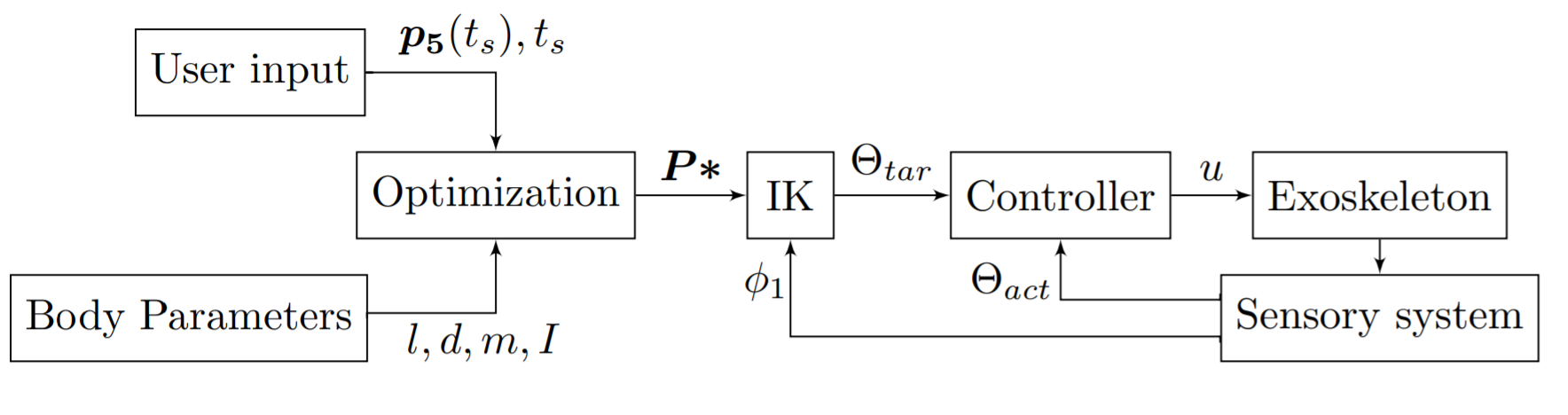}}

\begin{tikzpicture}
\end{tikzpicture}
\vspace{-4mm}
\caption{Block diagram of the system is shown to illustrate the connections between the modules.}
\vspace{-4mm}
\label{block}
\end{figure}
\begin{figure}[!htb]
\centerline{\includegraphics[scale = 0.28]{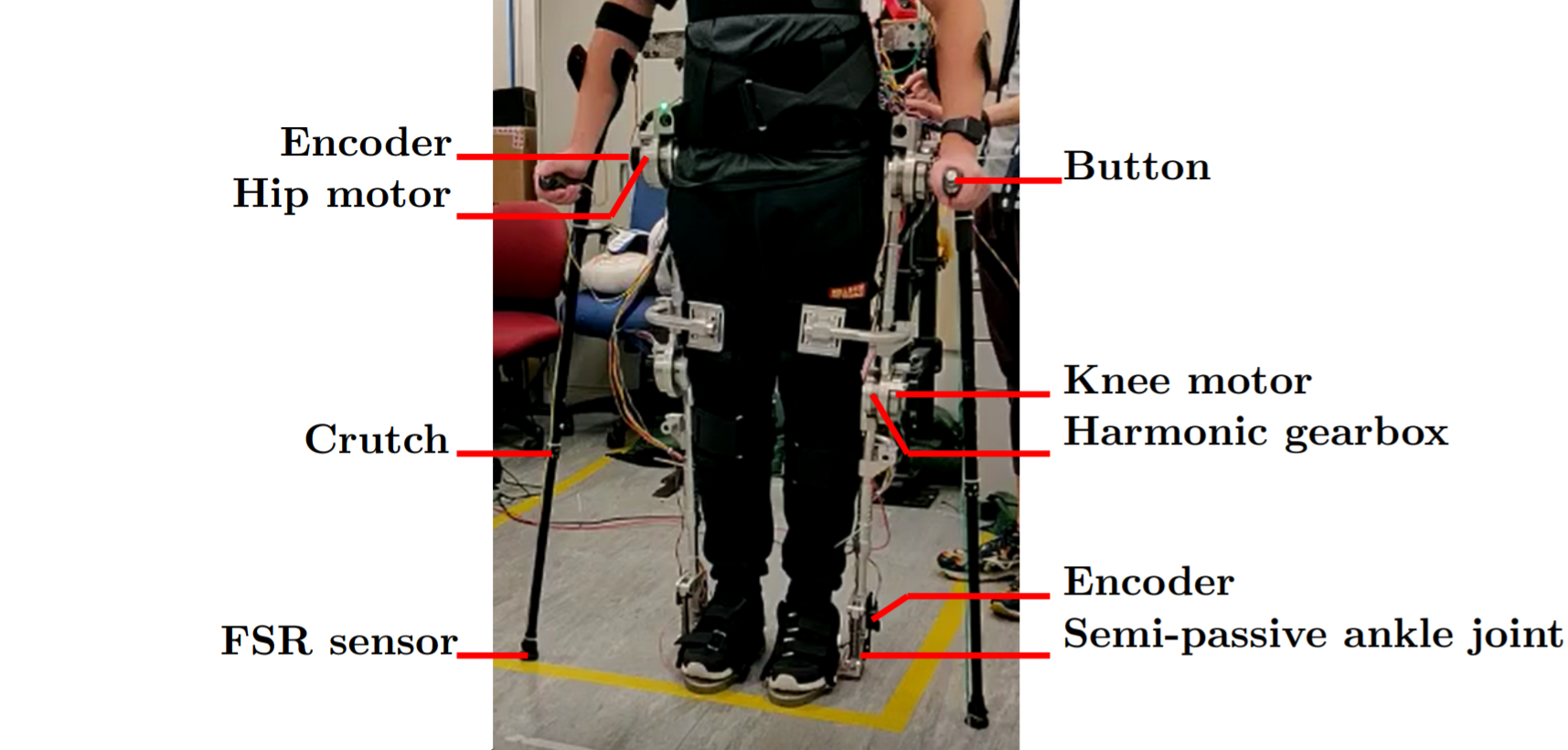}}

\begin{tikzpicture}
\end{tikzpicture}
\vspace{-4mm}
\caption{Exoskeleton test platform.}
\vspace{-4mm}
\label{fig}
\end{figure}

\begin{figure}[!htb]
\centerline{\includegraphics[scale = 0.20]{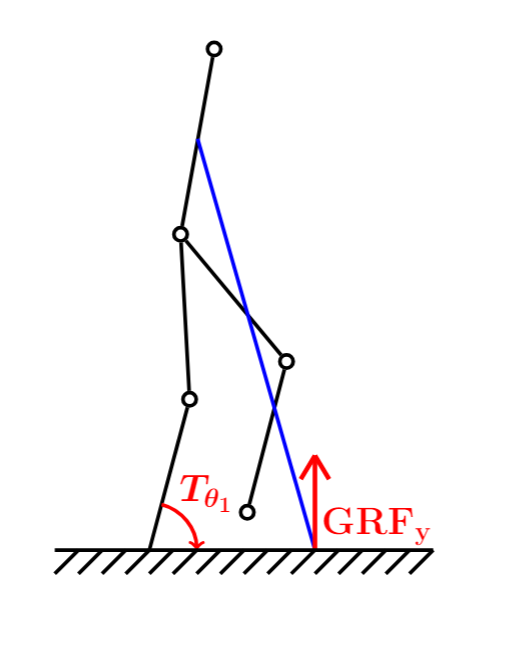}}
\vspace{-4mm}
\caption{The 5-link model of the exoskeleton with a blue crutch in the single support phase on sagittal plane. On the sagittal plane, the ankle torque of the 5-link model is proportional to the y-component of GRF during single support phase, so it is used for approximating the GRF in the optimization  stage.}
\vspace{-4mm}
\label{explain}
\end{figure}

\begin{figure}[t]
\centerline{\includegraphics[scale = 0.36]{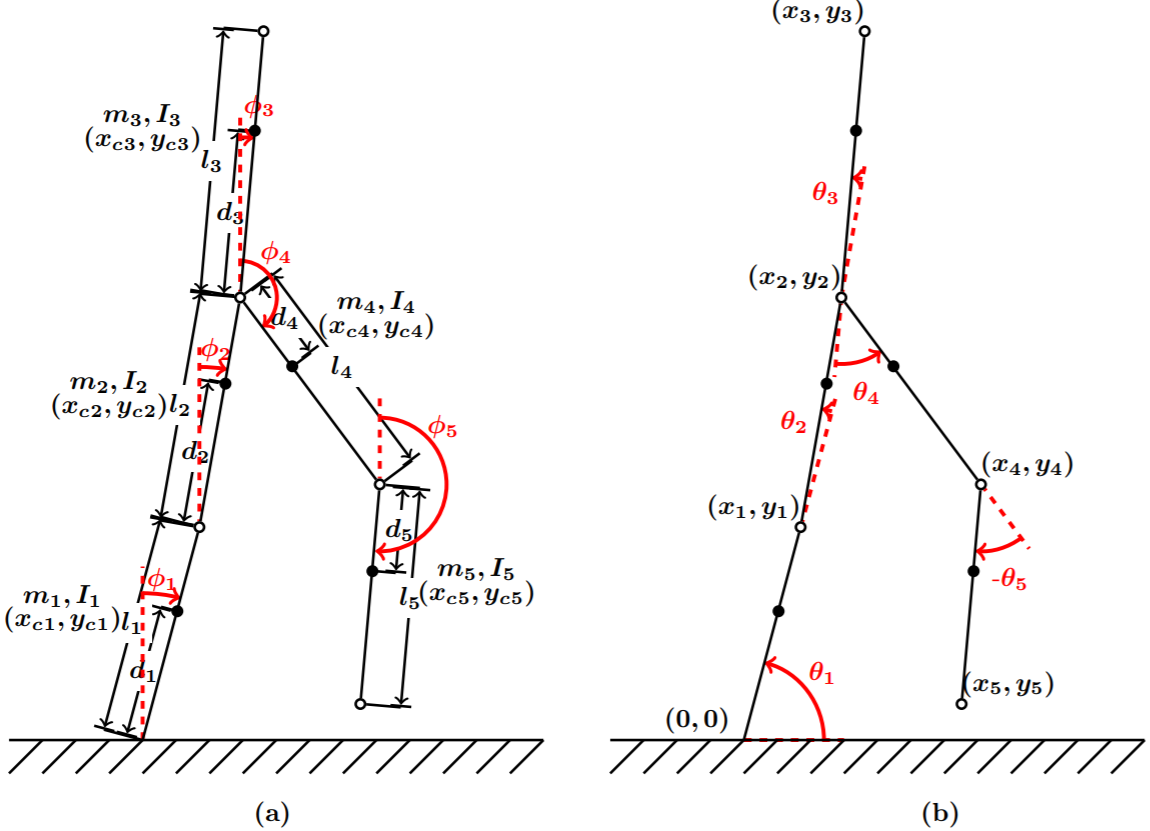}}
\vspace{-4mm}
\caption{5-link model of exoskeleton in swing {phase: }(a) in absolute angle system. (b) in relative angle system.}
\vspace{-4mm}
\label{fig_5}
\end{figure}

\section{Kinematic and dynamic model}
A 5-link model is established to analyze {the motion} and torque {of} each joint which {will} be used to formulate the optimization problem\cite{5-link}. Dynamical model and inverse kinematic (IK) will be presented on the sagittal plane with respect to the DoF of the exoskeleton.

\subsection{Dynamical model}
Absolute angles are used for torque computation in Lagrangian mechanics to simplify the expression \cite{Larg Model}, and relative angle is used in motor position control. Both angle systems are explained and shown in Fig. \ref{fig_5}. The 5-link model {consists} of single support phase and instantaneous double support {phase. Only} swing phase is considered in the optimization problem because the pilot is only required to balance during the swing phase in sagittal plane.

The model parameters in Fig. \ref{fig_5} are defined as follows. $(x_i,y_i)$ is the coordinate of the end point of link $i$ in the sagittal plane($i$ = 1,2,\dots,5); $m_i$ is the mass of link $i$; $I_i$ is the moment of inertia of link $i$; $(x_{c_i},y_{c_i})$ is the coordinate of the center of mass (COM) of link $i$ and $d_i$ the distance between the COM of link $i$ and joint $i$; $\phi_i$ is the absolute angle of link $i$  defined as the clockwise angle between link $i$ and the vertical upward direction; $\theta_i$ is the relative angle of link $i$ which is defined as the angle between link $i$ and link $i-1$ (ground when $i-1=0$, and positive in anti-clockwise direction). Define vector $\bm{\Phi} = [\phi_1,\dots,\phi_5]^T$, $\bm{\Theta} = [\theta_1,\dots,\theta_5]^T$, $\bm{p_i}=[x_i,y_i]^T$and $\bm{A}$ be the transformation matrix from relative angle to absolute angle as shown in \eqref{lin_trans}.    

\begin{equation}\footnotesize
\label{lin_trans}
\setlength\arraycolsep{2pt}
\begin{gathered}
    \bm{A} = \begin{bmatrix} -1 & 0 & 0 & 0 & 0 \\
    -1 & -1 & 0 & 0 & 0 \\
    -1 & -1 & -1 & 0 & 0\\
    -1 & -1 & -1 & -1 & 0\\
    -1 & -1 & -1 & -1 & -1 \\
    \end{bmatrix},\bm{b} = \begin{bmatrix} \frac{\pi}{2} \\
    \frac{\pi}{2}\\
    \frac{\pi}{2}\\
    \frac{3\pi}{2}\\
    \frac{3\pi}{2}\\
    \end{bmatrix}\\
    \Phi = \bm{A}\Theta + \bm{b}
\end{gathered}
\end{equation}

The dynamical model is established through Lagrangian dynamic equation which is defined for generalized coordinate system and it can be expressed as the equation in \eqref{larg}:

\begin{equation}\footnotesize
\label{larg}
\begin{gathered}
    L = K - P\\
    T_{q_i} = \dv{}{t}\pdv{L}{\dot{q_i}}-\pdv{L}{q_i}
\end{gathered}
\end{equation}

where $K$ is the kinetic energy of the system, $P$ is the potential energy of the system, and $L$ is Lagrangian which is the difference between $K$ and $P$. By taking the derivative, $T_i$ can be obtained which is the force in the generalized coordinate system $q_i$. Therefore, Lagrangian mechanics is dependent on the coordinate system { which will in turn} affect the result. The torque of the 5-link model in absolute angle can be expressed in \eqref{solution} where $\bm{D}(\Phi)\in \mathbb{R}^{5\times5}, \bm{H}(\Phi,\dot{\Phi})\in\mathbb{R}^{5\times5}$ and $\bm{G}(\Phi) \in \mathbb{R}^{5\times1}$.

\begin{equation}\footnotesize
    \label{solution}
    \bm{T}_\Phi = \bm{D}(\Phi)\ddot{\Phi}+\bm{H}(\Phi,\dot{\Phi})\dot{\Phi}+\bm{G}(\Phi) 
\end{equation}

\begin{equation}\footnotesize
    \label{sol1}
    \begin{gathered}
    D_{ij} = p_{ij}\cos(\phi_i-\phi_j) \\
    H_{ij} = p_{ij}\sin(\phi_i-\phi_j)\dot{\phi_j} \\
    G_i = -g_i\sin(\phi_i)
    \end{gathered}
\end{equation}

\begin{equation}\footnotesize
    p_{ij} =  \begin{cases}
    I_i + m_id_i^2 + a_i l_i^2 \left (\sum\limits_{k=1+i}^{5}m_k\right)  & j=i\\
    a_im_jd_jl_i + a_i a_j l_il_j \left (\sum\limits_{k=1+i}^{5}m_k\right) & j>i\\
    p_{ji} & j<i
    \end{cases}
\end{equation}

\begin{equation}\footnotesize
    g_i = m_id_i g +a_i l_i g\left (\sum\limits_{k=1+i}^{5}m_k\right)
\end{equation}

\begin{equation}
\footnotesize
    \label{sol2}
    a_i = \begin{cases}
        0 & i=3 \\
        1 & i=1,2,4,5
    \end{cases}
\end{equation}

Equations \eqref{sol1} - \eqref{sol2} are provided with details in \cite{Larg Model}. With the aid of absolute angle system, a relatively less complex solution is obtained but the torque expressed in absolute angle is not the joint torque. Instead it is the net torque of each joint. As a result, an additional step is required to convert the solution in absolute angle to relative angle with chain rule. Equation \eqref{lin_trans} shows that the absolute angle is a linear transformation of relative angle, so the relationship between partial derivative of $\phi$ verse $\theta$ is \eqref{haha}. The final step is to sum up all the derivative in chain rule to obtain the result of the joint torque \eqref{mid}, which can be expressed as \eqref{hehe} in matrix form, where $T_{\theta_1}$ is the ankle torque and it will { be used} for minimization in $\bm{T_\Theta} = [T_{\theta_1},T_{\theta_2},T_{\theta_3},T_{\theta_4},T_{\theta_5}]^T$. 
\begin{figure}[t]
\hspace{1cm}
\includegraphics[scale = 0.35]{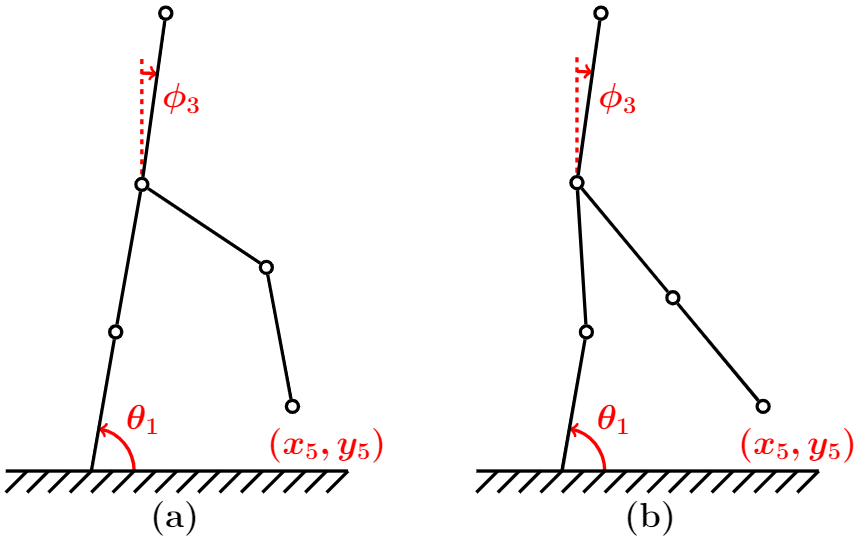}
\vspace{-2.5mm}
\caption{5-link model with fixed $(x_5,y_5)$,$\phi_1$ and $\phi_3$.(a) $r_2=1$. (b) $r_2=0$.}
\vspace{-2.5mm}
\label{fig3}
\end{figure}
\begin{equation}
\footnotesize
\label{haha}
\pdv{\phi_i}{\theta_j} = \pdv{\dot{\phi_i}}{\dot{\theta_j}} = A_{ij}\end{equation}

\begin{equation}
\footnotesize
    \label{mid}
    \pdv{L}{\theta_j} = \sum\limits_{i=1}^{5}\pdv{L}{\phi_i}\pdv{\phi_i}{\theta_j}
\end{equation}

{
\begin{equation}
\footnotesize
\label{hehe}
T_\Theta  = \bm{A^T}T_\Phi
\end{equation} 
}

\subsection{Inverse kinematic}
In this paper, $(x_5,y_5,\phi_1,\phi_3,r_2)$ are chosen to describe the gait trajectory instead of using joint angles $\theta_i$ {as in} most researches. IK is established to convert the variables chosen above to $\theta_i$ for both analysis and torque computation. Preliminary experiment is conducted with captured human gait data in \cite{gait2} with {exoskeleton, where two} main problems were encountered. The first problem is that the pilot required extensive force to balance, which has been {mentioned in} \cite{b6}. The second problem is the landing point of the swing leg at the end of swing phase is inconsistent due to the uncontrollable ankle joint. The ankle joint of test platform is spring-driven, so the pilot might land earlier or later than the landing point of the measured gait. This {can} cause instability and inaccuracy in stride length which is dangerous in stair climbing or obstacle avoidance.

Therefore, the end point trajectory $(x_5,y_5)$ of swing leg is used to describe the gait instead of the joint angles. {As the test platform} has 4 active DoFs, so two more parameters are needed to obtain a unique solution of the joint angle. In order to roughly control the COM of link 3 which is the heaviest part of the 5-link model, absolute angle of link 3, $\phi_3$, and ratio of range of relative angle of link 2, $r_2$, are used to describe the gait which is illustrated in Fig. 5. The reason of using $r_2 \in [0,1]$ instead of $\phi_2$ is because whether $(x_5,y_5)$ can reach the target is more important than the exact location of the COM of torso. There {exists} a range of $\phi_2$ that allows $(x_5,y_5)$ to reach the target, and $r_2$ describes {how and where} to choose $\phi_2$ from that range. Fig. \ref{fig3} demonstrates the effect of $r_2$ on fixed ankle joint angle $\phi_1$. Let the upper limit of $\phi_2$ be $\phi_{max}$  and lower {limit be} $\phi_{min}$. The steps to find all the absolute angles given $(x_5,y_5)$, $\phi_1$, $\phi_3$ and $r_2$ are shown below.

Equation \eqref{dfsadf} is to obtain absolute angle from consecutive joint positions and \eqref{lll} is to obtain the position of joint 1 from $\phi_1$ in vector form. 

\begin{equation}
\footnotesize
    \label{dfsadf}
    \phi_i = \frac{\pi}{2} - atan(\bm{p_i}-\bm{p_{i-1}})
\end{equation}

\begin{equation}
\footnotesize
    \label{lll}
    \bm{p_1} = \begin{bmatrix} l_1\sin\phi_1 \\ l_1\cos\phi_1 \end{bmatrix}
\end{equation}

 Equation \eqref{qw} and \eqref{dsfgf} are used to solve two circle intersection and only  the feasible solution is considered. It can be visualized as a circle with radius $l_2$ and center at $(0,0)$ and another with radius $l_4+l_5$ with center $(\|\bm{p_5}-\bm{p_1}\|,0)$. The intersection point is the farthest $\bm{p_2}$ away from target but still can reach the target $(x_5,y_5)$. {Therefore} $\frac{\pi}{2}-\psi(\phi_1,\bm{p_5})$ will the minimum or maximum $\phi_2$ that can reach the target.

\begin{equation}
    \label{qw}
    \setlength\arraycolsep{2pt}
    \footnotesize
    \psi(\phi_1,\bm{p_5})=\atan(\begin{bmatrix}
    B_1&-B_2\\B2&B_1
    \end{bmatrix}\frac{\bm{p_5}-\bm{p_1}}{\|\bm{p_5}-\bm{p_1}\|})
\end{equation}

\begin{equation}
\footnotesize
    \label{dsfgf}
    B_i= \begin{cases} \frac{l_2^2-(l_4+l_5)^2+\|\bm{p_5}-\bm{p_1}\|^2}{2\|\bm{p_5}-\bm{p_1}\|} & i=1 \\ \text{sgn}(x_5)\sqrt{l_2^2 - B_{1}^2} & i=2  
    \end{cases}
\end{equation}

By assuming that the target is always reachable, the only case {when} there is no solution in two circle interaction {will be that} the centers of the two circles are too close. As a result, a condition is designed to prevent no solution in  \eqref{qw} and \eqref{dsfgf}. The reason of having max and min function in \eqref{aaa} and \eqref{aba} is because certain $\frac{\pi}{2}-\psi(\phi_1,\bm{p_5})$ can reach the target in the 5-link model but the human joint cannot operate at that range, so the max and min function will ensure that the range of $\phi_{max}$ and $\phi_{min}$ will be always within human range.

\begin{equation}
 \label{aaa}
  \footnotesize
 \phi_{2max} = \begin{cases} 
 \phi_1 & x_5 \geq 0 \\
 \min\{\frac{\pi}{2}-\psi(\phi_1,\bm{p_5}),\phi_1\} & x_5<0 \\
  \phi_1 & \|\bm{p_5}-\bm{p_1}\|+l_2 \leq l_5+l_4
 
 \end{cases}
\end{equation} 

\begin{equation}
 \label{aba}
  \footnotesize
 \phi_{2min} = \begin{cases} 
 \phi_1-\frac{\pi}{4} & x_5 \leq 0 \\
 \max\{\frac{\pi}{2}-\psi(\phi_1,\bm{p_5}),\phi_1-\frac{\pi}{4}\} & x_5>0 \\
 \phi_1-\frac{\pi}{4} & \|\bm{p_5}-\bm{p_1}\|+l_2 \leq l_5+l_4
 
 \end{cases}
\end{equation}

\begin{equation}
\footnotesize
    \label{diiii}
    \phi_2 = r_2\phi_{2max} + (1-r_2)\phi_{2min}
\end{equation}

After obtaining $\phi_2$, $\bm{p_2}$ can be computed by \eqref{llll} and $\bm{p_4}$ can use two circle intersection to obtain the solution similar to \eqref{qw}. Absolute angle of all joints can be computed by \eqref{dfsadf} after all joint positions are obtained for both torque analysis and motor control. Therefore, with the aid of IK and dynamical model, the relationship between the optimized parameter $(x_5,y_5)$, $r_2$ and $\phi_3$ and torque of ankle joint {$T_{\theta_1}$} can be established which can be used in the optimization in the next section.

\begin{equation}
\footnotesize
    \label{llll}
    \bm{p_2} = \begin{bmatrix} l_1\sin\phi_1+l_2\sin\phi_2 \\ l_1\cos\phi_1+l_2\cos\phi_2 \end{bmatrix}
\end{equation}

\begin{equation}
\footnotesize
    \label{leell}
    \setlength\arraycolsep{2pt}
    \bm{p_4} = \begin{bmatrix}
    C_1 & -C_2 \\ C_2 & C1
    \end{bmatrix}\frac{\bm{p_5}-\bm{p_2}}{\|\bm{p_5}-\bm{p_2}\|}+\bm{p_2} 
\end{equation}

\begin{equation}
\footnotesize
    \label{dssgf}
    C_i= \begin{cases} \frac{l_4^2-l_5^2+\|\bm{p_5}-\bm{p_2}\|^2}{2\|\bm{p_5}-\bm{p_2}\|} & i=1 \\ \sqrt{l_4^2 - C_{1}^2} & i=2  
    \end{cases}
\end{equation}

\section{Ankle torque minimization}
\subsection{Gait description}

The goal of the entire optimization is to obtain the trajectory of $(x_5,y_5)$, $r_2$ and $\phi_3$ which minimize the ankle torque to reduce the load from the user upper body. As a result, the description of the trajectory based on the optimized parameter needs to be introduced. Bézier curve is used for describing the {trajectory whose} equations are shown in \eqref{Brcurve} and \eqref{Brcurve2} . There are two main reasons for using the Bézier curves instead of using polynomial directly{.} First, the curve will always go through the first control point and last control point. This ensures the starting position of curve to be the initial condition of the exoskeleton state {and prevents} any sudden shape change, while the last control point will allow the user to choose the precise landing point of the next step for $(x_5,y_5)$. The second reason is that the Bézier curve will always stay in the control polygon which reduces the difficulty for setting the optimization constraints to bound the trajectories. The formulations of Bézier curve of $r_2$ and $\phi_3$ are identical, so only the derivation of $r_2$ is shown in \eqref{Brcurve}. The $i$-th control points of Bézier curve of $r_2$ are denoted as $\bm{P^r_i} \in \mathbb{R}^{1\times2} $ where $i$ = 0,\dots,4 and $\bm{P^r} = [\bm{P^r_0}, \bm{P^r_1} ,\bm{P^r_2} ,\bm{P^r_3},\bm{P^r_4}]^T$. 

\begin{equation}
    \footnotesize
    \label{Brcurve}
    \setlength\arraycolsep{2pt}
    \begin{bmatrix}
        t & r_2
    \end{bmatrix}
    = 
    \begin{bmatrix}
    1 & u & u^2  &u^3 &u^4
    \end{bmatrix}
    \begin{bmatrix}
        1 &  -4  &   6  &  -4  &   1 \\
        -4 &  12  & -12  &   4  &   0 \\
         6 & -12  &   6  &   0  &   0 \\
        -4 &   4  &   0  &   0  &   0 \\
     1 &   0  &   0  &   0  &   0 
   \end{bmatrix} \bm{P^r}
\end{equation}

The first column of $\bm{P^r}$ is the control point of $t$ which is time and the second column is for $r_2$, which {implies that time is expressed} as a polynomial of $u$ where $u \in [0,1]$. However, the relationship between $r_2$ and $u$ or $t$ and $u$ are not the focus. Rather the relationship between $r_2$ and $t$ is needed for both optimization and execution by the exoskeleton. {Therefore} a discrete set of point $u$ is used to compute both the $r_2$ and $t$. Given a $t$, $r_2$ can be computed by linear interpolation. If the resolution of the $u$ increases, then the error of the $r_2$ will be reduced. 

The trajectory of the end point $(x_5,y_5)$ is defined by a 2D fourth-order and 2D third-order Bézier curves. The $i$-th control point of $(x_5,y_5)$ is denoted as $\bm{P^p_i} \in \mathbb{R}^{1\times2} $ where $i$ = 0,\dots,4 and $\bm{P^p} = [\bm{P^p_0}, \bm{P^p_1} ,\bm{P^p_2} ,\bm{P^p_3},\bm{P^p_4}]^T$. The curve describes the spatial relationship between $x_5$ and $y_5$ which {is} useful for obstacle avoidance in the future. In order to determine the velocity and acceleration of $(x_5,y_5)$, another variable $z$ is introduced to control the pace which is a 2D third-order Bézier curve. The $i$-th control point of $z \in [0,1]$ is denoted as $\bm{P^z_i} \in \mathbb{R}^{1\times2} $ where $i$ = 0,\dots,3 and $\bm{P^z} = [\bm{P^z_0}, \bm{P^z_1} ,\bm{P^z_2} ,\bm{P^z_3}]^T$. The computation of $z$ from $t$ is identical to $r_2$ from $t$ in \eqref{Brcurve}. The relationship between $(x_5,y_5)$ and $z$ and $t$ is shown in \eqref{Brcurve2}.

\begin{equation}
    \footnotesize
    \label{Brcurve2}
    \setlength\arraycolsep{2pt}
    \begin{gathered}
    \begin{bmatrix}
        x_5 & y_5
    \end{bmatrix}
    = 
    \begin{bmatrix}
    1 & z & z^2  &z^3 &z^4
    \end{bmatrix}
    \begin{bmatrix}
        1 &  -4  &   6  &  -4  &   1 \\
        -4 &  12  & -12  &   4  &   0 \\
         6 & -12  &   6  &   0  &   0 \\
        -4 &   4  &   0  &   0  &   0 \\
     1 &   0  &   0  &   0  &   0 
   \end{bmatrix} \bm{P^p} \\
       \begin{bmatrix}
        t & z
    \end{bmatrix}
    = 
    \begin{bmatrix}
    1 & u & u^2  &u^3 
    \end{bmatrix}
    \begin{bmatrix} 
    -1  &   3  &  -3  &   1  \\
     3  &  -6  &   3  &   0  \\ 
    -3  &   3  &   0  &   0  \\
     1  &   0  &   0  &   0  
   \end{bmatrix}\bm{P^z}
   \end{gathered}
\end{equation}

After obtaining the trajectory of $x_5,y_5$, $r_2$ and $\phi_3$, $\bm{\Phi}$ should be computed through IK for torque computation for optimization. However, $\phi_1$ is controlled by the pilot which is not available during the optimization stage. In order to simplify the problem, the $\phi_1$ is assumed to be a logistic function in this stage with predefined parameter. With this assumption, $\bm{\Phi}$ can be obtained by IK. Since the derivatives are needed for torque computation, it can be approximated in \eqref{ee2} where $K$ is the time-stamp. {Now all} the tools needed to compute $T_{\theta_1}$ from $\bm{P^r}$, $\bm{P^\phi}$,$\bm{P^p}$, $\bm{P^z}$,$\phi_1$ and $k$ are presented. For clarity of presentation, let \eqref{sim_torque} be the function to compute the ankle torque with the parameters above replacing all the steps above, with $\bm{P} = [\bm{P^{rT}},\bm{P^{\phi T}},\bm{P^{pT}},\bm{P^{zT}}]^T $ . 
\begin{equation}
    \footnotesize
    \label{ee2}
    \begin{gathered}
    \bm{\dot{\Phi}}(k)K = \bm{\Phi}(k) - \bm{\Phi}(k-1)
    \\
    \bm{\ddot{\Phi}}(k)K = \bm{\dot{\Phi}}(k) - \bm{\dot{\Phi}}(k-1)
   \end{gathered}
\end{equation}

\begin{equation}
    \footnotesize
    \label{sim_torque}
    T_{\theta_1} = \uptau(\bm{P},\phi_1(k),k)
\end{equation}

\subsection{Optimization problem}
As mentioned before, the ankle torque will only be minimized when the leg is swinging, and the ankle torque in double-leg support phase will not be {considered}. When the swing leg is still in contact with the ground, the ankle torque of the standing leg represents the GRF of swing leg which {implies} that the swing leg is supporting the body. The goal of the optimization is to reduce the load from the pilot, so when the swing leg touches the ground, the pilot would not need to apply much force to balance the body. Minimizing this period of ankle torque would not reduce the load of the pilot. By ignoring ankle torque in that period, the GRF of the swing leg can be used to accelerate and decelerate the body in a way that the pilot load is minimized when the leg is swinging, not being limited by the optimization algorithm. Therefore, \eqref{dground} is {used} to determine whether the swing leg is touching the ground and will be used in the cost function: 0 implies having contact, 1 for no contact. Only the initial and landing points are needed to be considered, because the constraints of the control points will prevent the swing leg from touching the ground outside of those two points.{ The variables} $\bm{p_5(0)}$ and $\bm{p_5(t_s)}$ are defined as the coordinates of the starting and landing points of swing leg, {and} $t_s$ is the step time.

\begin{equation}
    \footnotesize
    \label{dground}
    c(\bm{p_5},\bm{p_5(0)},\bm{p_5(t_s)}) = 
    \begin{cases}  0 &  \|\bm{p_5}- \bm{p_5(0)}\| \leq 0.002   \\ 0 &\|\bm{p_5}-\bm{p_5(t_s)}\| \leq 0.002 \\ 1 &\mbox{otherwise} \end{cases}
\end{equation}


\begin{equation}\footnotesize\label{cons}
\setlength\arraycolsep{2pt}
\bm{B} = \begin{bmatrix}
0 & r_2(0) \\
0 & 0 \\
0 & 0 \\ 
0 & 0 \\ 
t_s & 0 \\ 
0 & \phi_3(0)\\
0 & -\frac{\pi}{12} \\
0 & -\frac{\pi}{12} \\ 
0 & -\frac{\pi}{12} \\ 
t_s & -\frac{\pi}{12} \\ 
x_5(0) & y_5(0) \\
x_5(0) & y_5(0)+300 \\
x_5(0) & \min\{y_5(0),y_5(t_s)\} \\
x_5(t_s) & y_5(t_s)+300 \\
x_5(t_s) & y_5(t_s) \\
0 & 0 \\
0 & 0 \\ 
0 & 0 \\ 
t_s & 1 \\
\end{bmatrix}, \bm{U} = \begin{bmatrix}
0 & r_2(0) \\
t_s & 1 \\
t_s & 1 \\ 
t_s & 1 \\ 
t_s & 1 \\ 
0 & \phi_3(0)\\
t_s & \frac{\pi}{12} \\
t_s & \frac{\pi}{12} \\ 
t_s & \frac{\pi}{12} \\ 
t_s & \frac{\pi}{12} \\ 
x_5(0) & y_5(0) \\
x_5(0) & 600 \\
x_5(t_s) & 600 \\
x_5(t_s) & 600 \\
x_5(t_s) & y_5(t_s) \\
0 & 0 \\
t_s & 1 \\ 
t_s & 1 \\ 
t_s & 1 \\
\end{bmatrix} 
\end{equation}

\begin{equation}\footnotesize\label{cost}
\begin{aligned}
J(\bm{P})= & \frac{1}{K}\left(\sum\limits_{k=0}^{N}(c(\bm{p_5(k)},\bm{p_5(t_0)},\bm{p_5(t_s)}))(\uptau(\bm{P},\phi_1(k),k))^2 \right) \\
& + v_p \max_k(\bm{\dot{p_5(k)}})
\end{aligned}
\end{equation}

\begin{equation}\footnotesize\label{argcost}
\bm{P^\star} = \argmin_{\bm{B \preceq P \preceq U}} J(\bm{P}) 
\end{equation}

The upper bound matrix $\bm{U}$ and lower bound matrix $\bm{B}$ is defined in \eqref{cons}, {where} $\preceq$ is defined as {element-wise} comparison.{ The element in $\bm{U}$ and $\bm{B}$ are used as the constraints of the  Bézier curves control {points to ensure that the control polygon} and the trajectory will stay in range}. In \eqref{cons}, there are some control points {where} the upper bound is equal to the lower bound, because those points are determined by the initial condition and ending condition. An additional constraint is used to prevent dragging the {swing. The} $x$ control point of $\bm{P^p_1}$ and $\bm{P^p_3}$ are set to be the same as initial point and landing {point,} respectively, and lower limit of $y$ of the control point is set to be at least a fixed amount above the initial point and landing point.

The optimal solution of $\bm{P}$ for cost function \eqref{cost} is denoted as $\bm{P^\star}$. The cost function {sums} up all the squares of ankle torque in the swinging stage and a penalty term $v_p \max_k(\bm{\dot{p}_5(k)})$ is used to limit the maximum velocity of the end point trajectory where $v_p$ is a weight. In the simulation, the resulting end point trajectory reaches the landing point with an impractically high velocity. This is due to {that} the cost of the deceleration of the swing leg {is} ignored if it is decelerated through the GRF of the leg. The optimization problem stated in \eqref{argcost} {is solved by the} optimization tool box in MATLAB.

\section{Results}
\subsection{Experiment setup}

The GRF of crutches will be used to evaluate the effectiveness of different {gaits. A} metric representing the forces, defined in \eqref{GRF}, is {introduced, where $F$ denotes the} final result used to compare different {gaits}, $f_{l}(k)$ and $f_{r}(k)$ are the GRF measured by the FSR under the left and right crutches in Newton at time $k$, and $T_{s}$ is the sampling time which is 0.01 second.

\begin{equation}\footnotesize\label{GRF}
   F  = T_{s} \sum\limits_{k=0}^{N} \left( f_{l}(k)+f_{r}(k) \right)
\end{equation}

{Two} sets of experiments were tested on the healthy subject, and {the lower body of pilot is relaxed during the process}. In the first set of experiment the reference joint angles were inputting directly to the exoskeleton using the data from \cite{gait2} for walking on flat ground. The second experiment {used} optimized result with IK to evaluate the effectiveness of the algorithm. 

\begin{table}[htbp]
\vspace{-4mm}
\caption{Model parameters of exoskeleton system (including pilot)}
\begin{center}
\begin{tabular}{lclclclcl}
\hline
& $l$(m) & $d$(m)  & $m$(kg) & $I$(kg m$^2$)\\ \hline
link 1   & 0.441 & 0.269 & 7.05 & 0.226\\ 
link 2   & 0.395 & 0.228&  10.5& 0.626 \\ 
link 3   & 0.714 & 0.342& 57.7 & 9.44\\ 
link 4   & 0.395 & 0.167 & 11.5 & 0.626\\ 
link 5   & 0.441 & 0.172 & 6.05 & 0.226\\ 
\vspace{-4mm}
\end{tabular}
\label{tab3}
\end{center}
\end{table}

The exoskeleton system including the pilot weighs 93.5 kg and scales 175 mm height. The pilot {was} asked to repeat the same gait 25 cycles which allowed the pilot to familiarize with the motion of the exoskeleton, and the trial cycles were fixed to prevent any performance difference due to more or less practice from specific gait. The pilot was asked to find the most comfortable crutch position in the test gait and maintain the same through out the entire experiment to reduce the difference in GRF caused by different crutch position. Then, the pilot was asked to perform the gait 10 more times. The GRF is recorded from the crutches; only the data obtained from the right swing {was used} for analysis. The uneven mass distribution between the left and right body and left and right hand preference might lead to different result measured from left swing leg and right swing leg. Therefore, all the experiment focused on the same leg. The stride lengths and step times depended on the reference joint angle obtained from \cite{gait2} and \cite{gait_stair} on flat ground and {stairs,} respectively. The proposed method used the same stride length and step time as \cite{gait2} and \cite{gait_stair} to ensure the fairness of the experiment. After computing the $F$ from all the trials, the maximum three and minimum three data set were removed to prevent outliers affecting the result.    

Because the leg length and leg ratio of the pilot might not be the same as the subjects in \cite{gait2} and \cite{gait_stair}, minor modifications were made in the amplitude of the joint angle to ensure { that the} gait enables the climbing on stairs. In the ground walking experiment, the stride length was 600mm and step time was 2.24s. In the stair climbing experiment, the height and depth of the stair were 93.5mm and 275mm, respectively, and the step time was 2.28s. The pilot stood with the left leg at the first step, had right leg at ground level, and ended with right leg on the second step. 

\begin{figure}[t]
\label{result}
\vspace{4mm}
\centerline{\includegraphics[scale = 0.28]{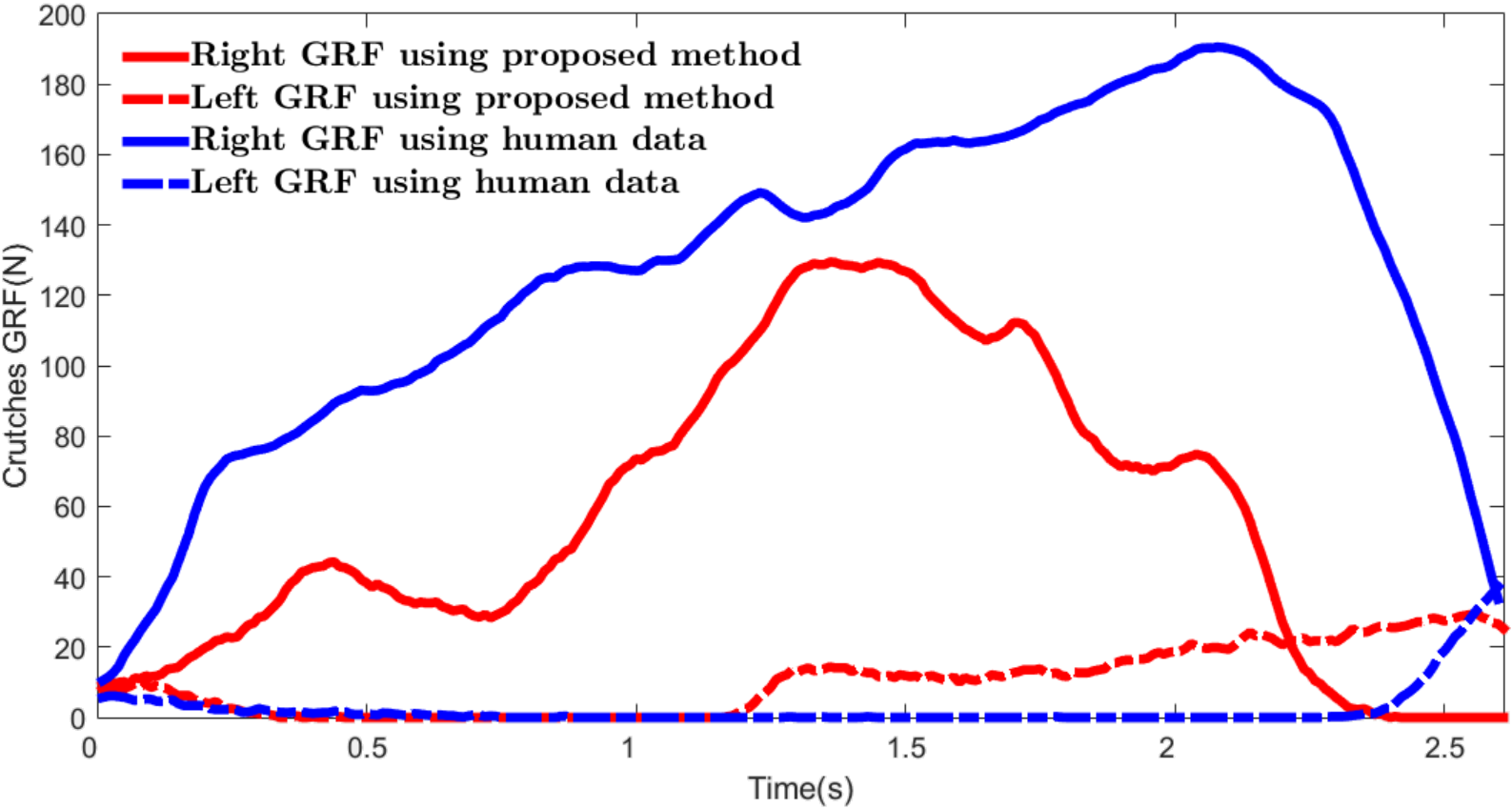}}
\vspace{-4mm}
\caption{Ground walking result plot}
\vspace{-4mm}
\end{figure}
\begin{figure}[t]
\label{result2}
\centerline{\includegraphics[scale = 0.28]{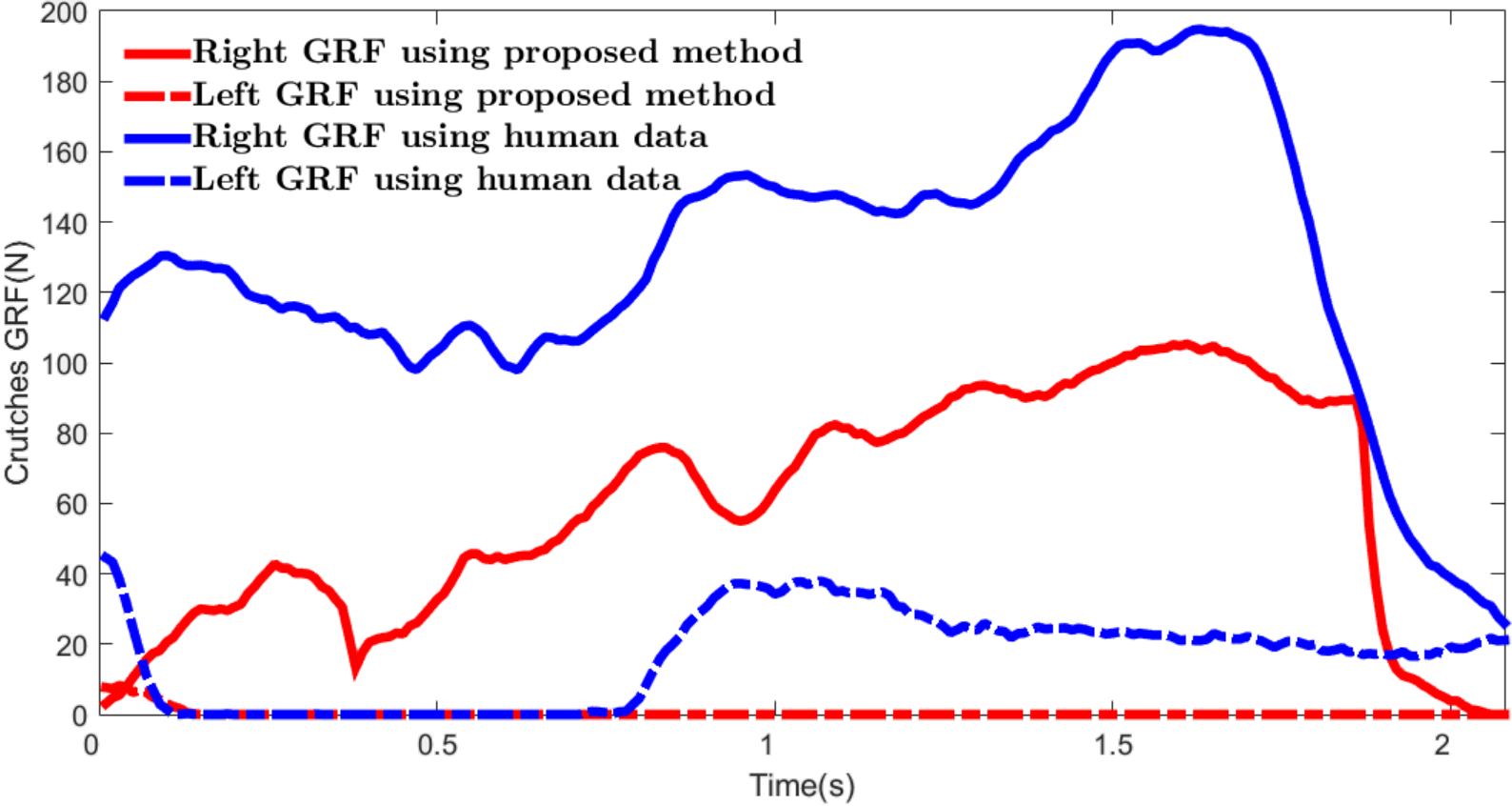}}
\vspace{-4mm}
\caption{Stair climbing result plot}

\end{figure}

\subsection{Experimental results}
Peak GRF and $F$ of each crutches were used for analysis; while the mean and standard deviation (SD) are shown in Table \ref{tab2}. A sample set of data from ground walking and stair climbing are plotted in Fig. 4 and Fig.5, respectively. Even though the step times of both experiments were around 2.2s, the pilot {was} required to balance the body through the crutches after the swing leg land, so the $F$ would keep incremental until the GRF dropped to close to 0. {Hence}, the time axis is beyond the time step.

\begin{table}[htbp]
\vspace{-4mm}
\caption{Table of experiment result}
\begin{center}
\scriptsize
\begin{tabular}{lclcl}
\hline
Ground walking   &$F$ &  left peak GRF & right peak GRF \\ \hline
 using human data & 230.21(18.83)  &  44.70(24.47) &  170.67(13.98) \\
 proposed method & 172.38(6.37)  &  19.17(9.05) &  138.18(6.20) \\ 
 reduction percentage & 25.1\% &  57.1\% & 19.0\% \\
\hline \\
 \hline
Stair climbing &$F$ &  left peak GRF & right peak GRF \\ \hline
 using human data & 405.64(19.69) &  54.36(19.83) &   179.87(24.88) \\
 proposed method &  102.71(20.83) &  11.37(3.09) &  81.20(16.11)  \\
 reduction percentage & 74.7\% &  79.1\% & 54.9\% \\\hline

\vspace{-4mm}
\end{tabular}
\label{tab2}
\end{center}
\end{table}

The reason why peak GRF is also considered in the experiment is that the goal is to reduce the load of the upper extremity of the pilot to prevent any potential damage on shoulder joint, and the fact {that even if} $F$ is minimized, peak GRF may still harm the pilot if it is too high. Therefore, the peak GRF was monitored in the experiment.

The ankle torque minimization reduced 25.1\% of $F$ {compared with} using human data in ground walking and 74.7\% in stair climbing, which validated the assumption of using ankle torque as an approximation of GRF from the pilot. Not only did it {reduce} $F$ according to the cost function \eqref{cost}, but also the peak GRF from the right crutch by 19.0\% and 54.9\% in flat ground walking and stair climbing, respectively. The reduction in peak GRF implies that the optimization will not create any high amplitude pulse in GRF to harm the pilot during the process.

There is an interesting observation about the experimental result. {The} summation of GRF $F$ from stair climbing is lower than the $F$ from ground walking from the proposed method. Pilot needs additional force to accelerate body vertically to achieve stair climbing, so $F$ from stair climbing is expected to be higher than ground walking. There is currently not enough evidence to {provide} a concrete explanation for this {phenomenon. More} experiment {need be conducted} to further enrich the knowledge about the GRF and ankle torque in the future.

\section{Conclusion}
In this paper, an ankle torque minimization approach is proposed in lower limb exoskeleton trajectory generation to reduce the upper extremity load and to provide precise end-point control of the swing leg. The dynamical model, IK model and cost function of the optimization problem were presented. The experimental result confirms the effectiveness of the proposed method by showing the significant reduction in GRF. It also shows that there is relationship between the ankle torque in 5-link model and crutches {GRF. More} researches could be done on exploring the model between GRF and ankle torque, and reducing GRF could be one of the {objectives} in exoskeleton gait trajectory generation in the future.

There are still some limitations in the current method. The offline optimization trajectory generation {requires} the ankle joint trajectory to compute the torque, but the ankle joint is uncontrollable by the exoskeleton. If the trajectory can be optimized according the actual joint angle in real-time, the performance of the system could be improved. Apart from the ankle joint limitation, the run-time of the algorithm is not ideal for  real-life application. The run-time of the method can be reduced by generating a large set of samples with different inputs and use supervised learning to obtain the input output relationship.

\end{document}